\RequirePackage{fix-cm}
\documentclass[smallcondensed]{svjour3}     
\smartqed  
\usepackage{graphicx}
\usepackage{paralist}
\usepackage{amsmath,amssymb,amsfonts}
\usepackage{cite}
\usepackage{url}
\begin{document}

\title{Playing a Strategy Game with Knowledge-Based Reinforcement Learning
}

\author{Viktor Voss         \and
        Liudmyla Nechepurenko \and
        Dr. Rudi Schaefer \and
        Steffen Bauer
}


\institute{V. Voss \at arago GmbH,
              \email{vvoss@arago.co}          
            \and
           L. Nechepurenko \at arago GmbH,
              \email{lnechepurenko@arago.co}
           \and
           Dr. R. Schaefer \at arago GmbH,
              \email{rschaefer@arago.co}
           \and
           S. Bauer \at arago GmbH,
              \email{sbauer@arago.co}
           \and
}

\date{Received: date / Accepted: date}

\maketitle

\begin{abstract}
This paper presents Knowledge-Based Reinforcement Learning (KB-RL) as a method that combines a knowledge-based approach and a reinforcement learning (RL) technique into one method for intelligent problem solving.  
The proposed approach focuses on multi-expert knowledge acquisition, with the reinforcement learning being applied as a conflict resolution strategy aimed at integrating the knowledge of multiple exerts into one knowledge base. 

The article describes the KB-RL approach in detail and applies the reported method to one of the most challenging problems of current Artificial Intelligence (AI) research, namely playing a strategy game. The results show that the KB-RL system is able to play and complete the full FreeCiv game, and to win against the computer players in various game settings. Moreover, with more games played, the system improves the gameplay by shortening the number of rounds that it takes to win the game. 

Overall, the reported experiment supports the idea that, based on human knowledge and empowered by reinforcement learning, the KB-RL system can deliver a strong solution to the complex, multi-strategic problems, and, mainly, to improve the solution with increased experience.

\keywords{Knowledge-Based Systems \and Reinforcement Learning \and Multi-Expert Knowledge Base \and Conflict Resolution}

\end{abstract}

\section{Introduction}
\label{intro}
Knowledge-Based Systems (KBS) have a long history in the field of AI. Even though they lost popularity in the last few decades, they have successful practical applications in many areas \cite{KBSrevevaluation,oravec,empiricalstudy}. KBS make use of human knowledge and experience to automate problems by means of machine reasoning. They were widely researched throughout the 1950's - 1990's and many areas of AI and Intelligent Applications emerged from this field \cite{empiricalstudy}. 

More recently, Machine Learning (ML) gained close attention and widespread acceptance among scientists, scholars and engineers as a promising technique for AI. Sub-fields such as Neural Networks, Reinforcement Learning, and Generative Adversarial Networks solve previously impossible problems and are very actively researched. It is no surprise that many studies investigate the possibility to fuse ML with other AI approaches aiming to achieve new breakthroughs. By combining different approaches, one attempts to overcome the challenges of a particular technique and benefit from the mutual advantages of various methods.

The proposed KB-RL method incorporates reinforcement learning into the knowledge-based system to handle conflicting or redundant knowledge. This allows the KBS to adopt multiple solutions for the same problem from multiple experts. The requirement of unambiguous knowledge imposes a restriction on the KBS and results in a costly human effort during knowledge engineering. Empowered by RL, the conflict resolution process can be automated and used to optimize the problem solution based on the experience gained by the KBS in solving such problems.

Playing various types of games is a common benchmark in Artificial Intelligence research. After AI systems beat the best human players at Chess \cite{chess}, Atari \cite{atari}, and then Go \cite{masteringGo}, strategy games have become the next level of complexity to challenge AI research. In 2019 DeepMind's AlphaStar demonstrated a compelling performance, outplaying humans in the real-time strategy game StarCraft \cite{alphaStar}. The winning technique was Deep Reinforcement Learning which fused Deep Neural Networks and Reinforcement Learning, allowing the model to learn from the large amount of games.

 We use the strategy game CIVILIZATION to showcase the capabilities of our KB-RL method. Particularly, the open-source version of the game FreeCiv was played by KB-RL agents. In addition to the complexity of the FreeCiv game, we were motivated by the proximity of the game paradigm to real world challenges.

\section{Related Work}

Nowadays, the field of KBS is highly heterogeneous and lacks coherent structure and clear formalism \cite{empiricalstudy}. The most researched sub-domains of KBS are Case-Based Reasoning, Fuzzy Systems, Multi-Agent Systems, Decision Support Systems, Cognitive Systems and Intelligent Software Agents \cite{empiricalstudy}. With the rise of Machine Learning methods, such as Neural Networks and Reinforcement Learning, it is not surprising that these techniques have been considered in application to KBS. 

Considering the diversity of research on KBS, it is difficult to compare our work to other studies on KBS in combination with RL, especially in application to broad problems such as managing complex environments like strategy games. 
Many researchers in the field of KBS used strategy games, and particularly FreeCiv, as a benchmark for their AI approaches. However, most of them focused on the specific elements of the game rather than playing the entire game. For instance, J. Jones, A. Goel, P. Ulam and their colleagues proposed a model-based reflection for self-adaptation for guiding reinforcement learning in the series of publications \cite{Jones2009MetareasoningFA,jones2009,jones2005,jones2005_2}. The authors demonstrated their methodology on the sub-elements of the FreeCiv game, such as building and defending cities. T.R. Hinrichs and K.D. Forbus studied how structural analogy in combination with qualitative reasoning can improve the prediction of population growth in FreeCiv civilization \cite{Hinrichs2,Hinrichs1}. 
Outside of KBS, works \cite{GAandFreeciv2} and \cite{GAandFreeciv1} explored the utilization of Genetic Algorithms for the optimization of city placement and city development in FreeCiv. 

For learning to win in the FreeCiv game, Branavan et al. \cite{WinByReadingManuals} employed a Monte-Carlo framework for analyzing the text manuals. Their work involves Natural Language Processing for text analysis and Reinforcement Learning for training the agent. As a result, the language-aware agent showed a significantly increased win rate (27\% to 78\%) in contrast to the agent not supported by the linguistic analysis. The games were played on a 36x24 map against one built-in computer player with the 'NORMAL' level of difficulty. Such game settings allowed to finish the game in less than 100 rounds, facilitating efficient reinforcement learning. Our research is different to Branavan et al. in both the methodology and the level of the game complexity.

Another endeavour to learn winning strategy games was made by M. Molineaux, D.W. Aha, and M. Ponsen \cite{Molineaux2012DefeatingNO,LearningToWin}. The authors employed Case-Based Reasoning to learn winning the real-time strategy game Wargus. They proposed a Case-Based Tactician system which learned to choose the best tactics out of several, utilizing three different sources of domain knowledge: state lattice, set of tactics and state cases. 

The most remarkable achievement in playing strategy games is the AlphaStar project \cite{alphaStar}. Based on Deep Reinforcement Learning, the AlphaStar software successfully won against two high ranked StarCraft II players. 

The aim of this article is to address and describe the KB-RL approach in its present condition. The reported experiment shows that the KB-RL method can be used successfully for large-scaled and sophisticated problems, such as playing strategy games by leveraging human heuristic knowledge and intelligent computer algorithms to reinforce learning.

\section{KB-RL method} \label{KBRLmethod}

\subsection{Knowledge-based system}

A Knowledge-Based System is a software system that contains a substantial amount of knowledge in an explicit, declarative form that is employed to reason about the problem matter \cite{KBSdefinition}. In contrast to conventional software programs, KBS do not embed the knowledge as part of the program code. Instead, the knowledge is captured in small fragments of human expertise, data, and information about the problem domain. Hence knowledge is manageable in a flexible way without the need to change and rebuild the system \cite{empiricalstudy}.

There are two main components that are expectedly present in knowledge-based systems: a knowledge base that accommodates the domain-specific knowledge and the problem-solving method (inference engine) which consists of algorithms for manipulating the knowledge to solve the presented problem \cite{expertsystemsbook}.

\subsection{Knowledge base}

The knowledge base contains two logically distinct components. One is factual knowledge that describes the environment of the problem, holds concepts, their properties, attributes and relationships. The second component is procedural (or inferential, or casual) knowledge that represents the heuristic knowledge and the expertise of human experts. In a rule-based KBS, inferential knowledge is presented in the form of rules that have to follow a specific syntax called knowledge-representation language \cite{businessinfosystems}. The KB-RL system described in this paper follows a rule-based approach, described in detail below.

The factual knowledge of the described approach is modelled as a semantic network. A semantic network is a graph-based knowledge formalism that provides a structural representation of concepts, their properties and relationships in the domain of interest \cite{semanticnetworks}. These concepts are modelled as nodes where node attributes hold the properties of the concept and the relationships are the arcs between the nodes.

The main benefit of semantic networks for knowledge representation is the possibility to translate arbitrary and unstructured information of human knowledge to a structured format that can be processed by the machine \cite{semanticnetworks}. Moreover, semantic networks enable the model to carry the semantics of the modelled world. This fact 
supports the knowledge-based system in reasoning about the knowledge and helps to establish common understanding of the data between computers and humans  \cite{KBSrevevaluation, sowa_semanticnetworks}. Furthermore, we chose a graph-based representation because in contrast to other data storage technologies (such as relational databases or NoSQL) it allows us to incorporate a necessary trade-off between the structure-first and the data-first approaches: on one hand, the data is semantically structured, on the other hand, it is still possible to store any type of information in the graph, as the knowledge is semi-structured.

\subsection{Ontology}

The design of the semantic network starts with defining an ontology. According to Gruber \cite{ontologydefinition}, an ontology is a formal specification of a shared conceptualization that owns high semantic expressiveness necessary for systems of increased complexity. Considering that the term "ontology" can be interpreted ambiguously \cite{ontologydefinition, ehrlinger2016towards, Guarino95ontologiesand}, we explicitly emphasize here that the ontology, as used in the described approach, is not equal to the knowledge base or knowledge graph. Rather, the role of the ontology is to define the schema for the semantic network in order to establish the common vocabulary and shared understanding of the data among people or software agents \cite{ontologydevelopment}. Having an agreement on terminology 
for all the concepts and their relationships facilitates  knowledge reuse and enables management of the information defined in the knowledge base \cite{ontoshare_an_ontology_based}. Though the ontology can also hold the class instances, in KB-RL the instances are held only in semantic network, and the ontology is a meta level for formalizing the semantic network structure.

The ontology of KB-RL follows the Resource Description Format \cite{RDF}, specifically the Turtle-Syntax \cite{Turtle}. Without going into detail, we should note that KB-RL's ontology is non-hierarchical, and does not imply inheritance. In general, all elements of the ontology are of one of the following types:
\begin{inparadesc}
\item[Entities] represent concepts (nodes in the graph),
\item[Verbs] are binary relations (edges) between two Entities and describe something an Entity does to or with another,
\item[Attributes] are properties of the Entity that hold a scalar value, such as a string or an integer, or a list of scalar values.
\end{inparadesc}
For more detailed information, see the Open Graph of IT \cite{OGIT}.

\subsection{Inference in KB-RL}

The process of deriving knowledge from a given knowledge base is known as inference \cite{expertsystemsbook}, and the problem-solving component of a knowledge-based system is therefore called an inference engine. 
Generally speaking, an inference engine acts as an interpreter that analyzes and processes the knowledge rules to derive a valid conclusion \cite{AkerkarKBSbook}.

Different KBS offer various types of inference, where most state-of-the-art systems employ an inference method based on the resolution principle \cite{expertsystemsbook}. In our KB-RL system, abductive reasoning \cite{abduction} is used to analyze the knowledge and derive the solution. Abductive reasoning seeks to form and evaluate the most likely hypothesis for the best possible explanation to the given problem based on the possibly incomplete evidences. One of the distinctive characteristics of abduction is the consideration of contextual knowledge in search of the solution \cite{expertsystemsbook}.
For example, the faults of a device can be diagnosed by finding a typical combination of conditions of the state of the device and of the expert knowledge about possible problems. The device state provides contextual information (which can be incomplete), and based on the available expert knowledge the device failure can be explained to the best match between the state conditions and expert knowledge. If the state changes, or new knowledge is available, the conclusion can change accordingly with the updated information.

\subsection{Issue entity}

\begin{figure}[t!]
\centering
\includegraphics[width=0.8\linewidth]{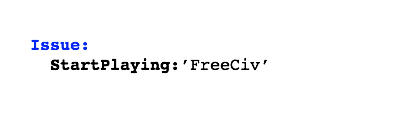}
\caption{An example of the Issue object as a holder for working memory in KB-RL system. The example shows the content of the Issue object on its creation. As more knowledge is applied to the problem, the Issue object will hold all relevant content in new attribute/value pairs. }
\label{fig:gameissue}
\end{figure}

In the KB-RL method, we distinguish between two kinds of contextual knowledge: one stems from the factual knowledge describing the environment and its state (e.g. the device is not responding), while the other characterizes the \textit{current} situation of the given problem (e.g. reset command was sent, waiting for response). In KBS theory it is called working memory \cite{expertsystemsbook}. For instance, this can be the information on what has already been done on the task, what is the goal of the task, or intermediate results in processing the task.

In our KB-RL method we call the working memory an Issue. The Issue is an object that represents a task and holds contextual information about it.
As an example, Figure \ref{fig:gameissue} shows an Issue object that is injected into the system to start playing the FreeCiv game. Originally it has only one attribute, \textit{StartPlaying}, with value 'FreeCiv', which instructs the inference engine to play the game. 
The Issue object is usually created from outside the system. Afterwards,  contextual information is only added and/or modified when the inference engine applies knowledge.

\subsection{Rules in our KB-RL system}

Procedural knowledge in the KB-RL system needs to be entered into the knowledge base by human experts in the form of rules. This process is called knowledge elicitation and usually involves two people: one with knowledge on the matter, and another person who is familiar with the knowledge-representation language and can encode the knowledge into the knowledge base. In the described experiment, we used a Protocol analysis \cite{elicitation} technique to acquire the knowledge from human experts.

In our KB-RL, the term Knowledge Item (KI) is used to refer to a single rule of the knowledge base. Hence, we will use the term Knowledge Item or KI interchangeably with the term 'rule'. The term KI derives from the principle that each rule is a single item of knowledge encoding an atomic action on the task. Essentially, we prefer to split the entire workflow on some task into granular steps in order to enable the reuse of the knowledge in similar but different tasks.

Technically, every Knowledge Item is a piece of code that contains the \textit{procedural knowledge} and the \textit{context} in which this knowledge is applicable. KIs are structured into four blocks:
\begin{itemize}
    \item[] \textbf{ki} - meta information about the rule,
    \item[] \textbf{on} - conditions on the factual context,
    \item[] \textbf{when} - conditions on the working memory,
    \item[] \textbf{do} - procedure to execute.
\end{itemize}

\begin{figure}[t!]
\centering
\includegraphics[width=1.0\linewidth]{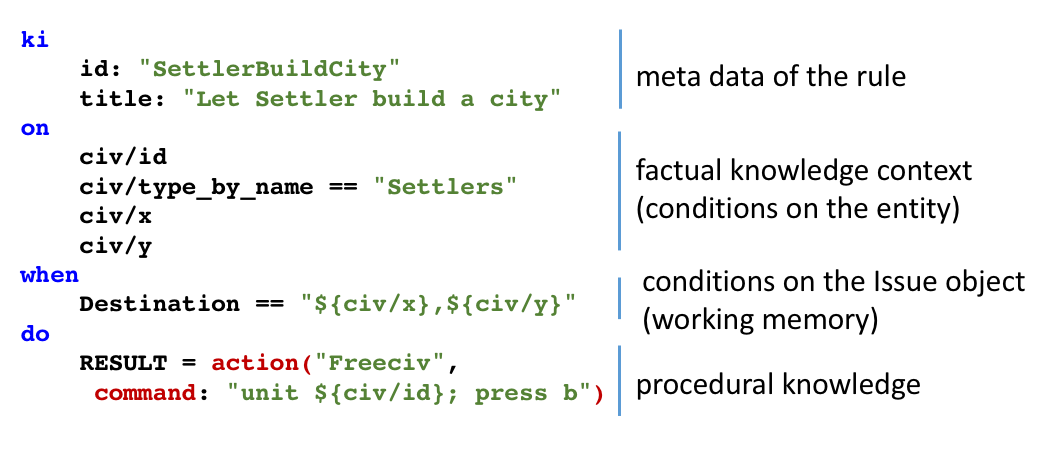}
\caption{An example of the Knowledge Item as a rule in KB-RL.}
\label{fig:kiexample}
\end{figure}

Figure \ref{fig:kiexample} shows the basic example of one Knowledge Item that performs building a city in the FreeCiv game. The ON condition specifies to which concept (an entity in the semantic graph) the knowledge is relevant. Technically, it defines which node the inference engine needs to find in the graph in order to execute the procedure given in the DO block. The WHEN block defines the condition on the context of the working memory (Issue object). When both ON and WHEN block conditions are met, the DO block will be executed. The example in Figure \ref{fig:kiexample} can be read like the following: if there is a node of type Settlers in the semantic graph, and it has attributes \textit{id}, \textit{x}, \textit{y}, ('civ/' prefix points to the namespace, we will ignore it for the sake of simplicity), and the Issue object has the attribute \textit{Destination} equal to the unit's coordinates, then execute the command given in the \textit{action()} function. The \textit{action()} function is named an Action Handler and is explained below. 

The example in Figure \ref{fig:kiexample} explains the main principle behind encoding the procedural knowledge into rules. Overall, KIs can be more complex than the given example, such as conditioning the relationships between nodes, making queries, manipulating lists, and others.

Together, all KIs constitute a knowledge base that the inference engine searches for suitable rules as it processes the Issue. When the KI is found to match the context conditions, it will be executed by the inference engine. As a result, the environment state and/or Issue object can change. For instance, after executing the KI given in Figure \ref{fig:kiexample}, the new city will appear in the game (and in the graph as a new node), and the Settler unit will disappear from the game and will be removed from the semantic graph. Consequently, the KI in Figure \ref{fig:kiexample} will become irrelevant while other KIs, for example KIs for city development, will become applicable to the new situation. In this fashion, the engine will apply KIs step by step to play the game until the game is finished.

\subsection{Connector}

While the KIs are entered into the system by knowledge engineers, the factual knowledge has to come from the environment that is managed by the system. For this purpose, a special software module is usually implemented that communicates with the environment and creates/updates the information in the graph via the REST API. We call this module the Connector. In the case of FreeCiv, the Connector was implemented as a Python program that ran parallel to the game and monitored the FreeCiv client for updates in the player's environment. With the any change in environment, e.g. when the turn ended or the unit moved, Connector sent updates to the corresponding nodes and arcs in the graph, allowing the engine to work with the up-to-date information.

\subsection{Action Handler}

In order to execute commands to external services, such as the FreeCiv client for example, our KB-RL system has a dedicated module named Action Handler (AH). An Action Handler is a highly configurable component that can perform actions, such as executing local or remote command line commands, running scripts, sending and receiving HTTP requests, communicating over websocket, and others. To send commands to a particular external service, the instance of the Action Handler has to be configured to specify the interface between the inference engine and the external service. For example, in the case of the FreeCiv game, the Action Handler "Freeciv" was configured to send commands to the FreeCiv client.

Without going into technical detail, the main idea behind the AH is that it is a way the KB-RL system interacts with the external environment and acts as an agent in respect to this environment. While the Connector only monitors the environment and updates the corresponding information in the KB-RL graph database, the Action Handler manipulates the environment and can change its state. For example, in the FreeCiv game it acts as a player the same way a human player would interact with the game.

\subsection{Multi-Expert Knowledge Acquisition}

The knowledge for a knowledge-based system can be acquired from one or multiple experts. Working with only one expert can make the knowledge acquisition process easier and smoother. However, obtaining the knowledge from a number of experts has certain advantages \cite{leary, delphi, easterbrook}. The group of experts can contribute to the improved quality of the knowledge base (its consistency, completeness, accuracy, and relevance),
achieving better productivity, addressing broader domains and more complex problems, and reducing the costs of knowledge access. Moreover, some problems cannot be solved by one expert but require the expertise of a team where each expert is highly knowledgeable about only a subset of the domain \cite{Mittal}.

On the other hand, multi-expert knowledge acquisition presents a serious challenge of integrating the knowledge of many into the knowledge base without contradiction and inconsistency. It is very likely that different people have different backgrounds and their own perspectives of the problem, use different terminologies for the same concepts, and have their own methods for solving the problem \cite{Mittal,shaw}. Therefore, eliciting knowledge from more than one expert can easily result in differing solutions for the problem, and consequently in alternative rules for it. Technically, it means that the inference engine will find more than one rule for the given problem context. This situation in KBS is called a knowledge \textit{conflict} \cite{Khalil}. Traditionally, conflicts are attributed to the disagreement in knowledge and understood as mistakes. KBS require the knowledge to be unambiguous for their inference engines to work \cite{Khalil}. Therefore, the conflict resolution process has to take place before the engine can execute the conflicting rules \cite{AkerkarKBSbook}.

\subsection{Conflict Resolution}

The conflict resolution can take place in different stages, such as during run time when the conflict directly occurs, or when inserting knowledge into the system, or during the knowledge elicitation phase. The simplest conflict resolution strategy at run time is based on the order in which the rules are found by the engine. For example, \textit{First in First Serve}, or \textit{Last in First Serve} \cite{Pakiarajah}. More advanced techniques involve context-sensitive criteria, for example, \textit{Prioritization}, \textit{Specificity}, \textit{Recency} \cite{expertsystemsbook}. While these methods can resolve conflicts automatically, the conflict resolution strategies in the knowledge engineering stage or in the knowledge elicitation stage mostly involve humans in the resolution process, for example \cite{clark, shaw, argumentation, katemes, leary, Mateu, statecharts}.

Conflict resolution is considered to be a restricting factor for KBS \cite{argumentation}, therefore, it has been studied intensively, especially in multi-expert knowledge-based systems where conflicts are more likely to occur. In addition to the simple conflict resolution techniques mentioned above, a number of more sophisticated methods have been proposed to support conflict resolution. Examples are an abductive device for conflict resolution \cite{menzies}, a multi-attribute support mechanism \cite{multiattribute}, incorporating expert's ranking \cite{temporal}, preferences and prioritized goals \cite{argumentation}, leveraging matrix representations and classification \cite{Mateu}, building statecharts for expert collaboration \cite{statecharts}, common conceptual model for conflict explanation \cite{shaw}, or employing multi-level hierarchical structures \cite{Puuronen}.

More recently, conflict resolution has been intensively studied in Multi-Agent Systems (MAS). The field of MAS grew out of the Knowledge-Based Systems umbrella \cite{empiricalstudy}, since the system that can perceive the environment and act in it is defined as an autonomous agent \cite{agentdefinition}. Furthermore, a multi-agent system is defined as a "loosely coupled network of problem-solving" agents \cite{agentdefinition}. Though conflict resolution in MAS is named the same as in KBS, they are not the same problem. In MAS, a conflict emerges between the agents that have conflicting interests. Thus, the agents can have a potentially fragmented view of the current system that may be inconsistent with the other agent's view, and may be outdated \cite{Khalil}. For the conflict resolution in MAS, it is critical to maintain communication between agents, to be able to negotiate or arbitrate on their goals.
On the contrary to MAS, the conflict in KBS appears in the knowledge rules for one agent that views the environment from a single perspective and does not need communication or negotiation between different entities. 

Traditionally, knowledge engineers seek to develop a consensus on conflicting rules and apply algorithms to single out one rule for the engine to proceed. However, such an approach in multi-expert systems potentially restricts the acquired knowledge due to filtering out the inconsistencies and introducing a consensual yet altered behavior of the experts \cite{easterbrook}. Though there is much research on dealing with multi-expert knowledge, little attention has been given to the possibility of preserving the knowledge of several experts in one knowledge base and turning conflicts into an advantage rather than seeing them as mistakes. Holding multiple solutions to one problem can be beneficial in many cases. For example, there are numerous marketing strategies that can be successful for a single business, or various process structures that are comparably effective in achieving the same organizational goals.

Let us consider a simplified example to demonstrate this idea. A computer hard drive often becomes full and requires freeing disk space. When asking different IT experts, one can suggest to first check the temporary files, while the other can point to the log files first. Both actions may solve the problem if tried. The recommendation depends on the expert's experience and may work for some systems and not be optimal for others. 

Projecting this elementary example to a more complex problem, such as playing FreeCiv for instance, there is more than one strategy to win the game. Moreover, the game was designed to be well balanced between different strategies \cite{balancedgame}. As such, boosting research and developing the economy can be as successful as aggressively fighting the opponents; building a small number of cities on limited territory can be as advantageous as settling on an entire continent. 

The examples above illustrate that on the scale of automating processes, such as business processes, where problems are complex and the environments are diversified, it can be beneficial to acquire knowledge from different experts and to employ diverse strategies within one system for the same task. From this perspective, traditional KBS present a limitation due to the requirement of consistent knowledge. In order to overcome this limitation, we insist that KBS need a refined conflict resolution approach that is capable of intelligent evaluation of the available knowledge and selecting the most advantageous knowledge for the assigned task.
We propose to employ reinforcement learning as a conflict resolution strategy in the knowledge-based system. We suggest that this technique enables multi-tactic solutions for various kinds of problems and allows learning an improved strategy for an assigned task. We have demonstrated our findings in the example of playing the empire-building multi-strategy game FreeCiv.

\subsection{Reinforcement Learning}

Reinforcement Learning is a ML approach that involves learning an agent's optimal behavior towards a predefined goal from the trial and error experience in the agent's environment. One of the most well-known sources of the RL definition and its detailed discussion can be found in \cite{Sutton1998}. Here we shortly note the main elements of RL.

RL is often defined as a Markov Decision Process (MDP) $<S,A,R,T>$ where $S$ is a set of possible states, and $A$ is a space of legal actions. Each state has a reward $R(s) \in  \mathbb{R}$ associated with it that can be implicitly provided by the environment. $T$ is often given as a transition distribution $p(s'|s,a)$ between states considering taken  actions. Here, $s'$ is the state following state $s$ after an action $a$ was executed. The transition distribution describes the dynamics of the environment and is called the \textit{model} of environment. The model in RL systems is an optional element that can be used, if available, for planning possible future situations before they are actually encountered. In contrast, model-free RL methods interact with the environment to learn about its dynamics.

The \textit{policy} $\pi$ is a mapping or distribution from state space to action space $S \rightarrow A$ that can be deterministic or stochastic. A stochastic policy can be described as a probability distribution of taking the action $a$ in state $s$ parameterized by an n-dimensional vector $\theta \in \mathbb{R}^n$, denoted as $\pi_\theta(a|s):S \rightarrow A$ \cite{bettaPolicy}. At each agent's step, a policy $\pi_0(s,a)$ is calculated from the distribution parameters, for example $\mu_\theta(s)$ and $\sigma_\theta(s)$ in Normal distribution.

Consider the agent with policy $\pi$ that starts from state $s_0$, chooses an action $a_0$, receives the reward $r_0 = R(s_0,a_0)$, then commutes to the next state $s_1$ and repeats this process. This will generate a sequence $\tau = s_0,a_0,s_1,a_1,...,$ that is called a trajectory of the agent. At some point in time, the agent will stop in some state $s_{end}$. This process of starting in state $s_0$ and arriving to the end state $s_{end}$ is called an episode. Each episode delivers a so-called $return$ of the episode that is denoted $G$ and defined as discounted cumulative reward over the episode: $G = r_0 + \gamma r_1 + \gamma^2 r_2 + ...$, where $\gamma$ is a discount factor in range between 0 and 1.

Running episodes one after another, the RL algorithm aims to learn an optimal policy $\pi^*$ that maximizes the expected return. To estimate the policy $\pi$ for a given state $s$, a state-value function  $V^\pi(s)=\mathbb{E}_\pi[r_0^\gamma|s_0=s]$ is defined as the expected return for the state $s$ when following the policy $\pi$. Alternatively, the action-value function $Q^\pi(s,a)$ can be used for learning the optimal policy $\pi$, where $Q^\pi(s,a)$ describes the value of the expected return starting from the state $s$, taking the action $a$, and following $\pi$ thereafter.

Many approaches in RL take advantage of the Bellman equation that expresses the recursive relationships between the value of a state and the values of
its successor states. Likewise, the Bellman optimally equation is generally used to derive the optimal policy from either the optimal value function or the optimal action-value function. For more detailed information refer to \cite{Sutton1998}.

\subsection{Monte Carlo Methods}

There are many kinds of RL algorithms for different types of RL problems, to name only a few of them: Dynamic Programming, Monte-Carlo Methods, Temporal Difference Learning, and others. All of them have their own flavor in terms of how they operate on the state-value or action-value functions, update policies, accumulate returns, optimize parameters, etc. For the KB-RL approach of the reported project, the Monte-Carlo Methods were used to learn the optimal policy for conflict resolution strategy. Monte Carlo methods are based on averaging over the sample returns. For the episodic tasks, the returns are averaged after every episode for each state visited in the episode. The idea behind Monte Carlo methods is that with more returns observed, the average should converge to the expected value.

The benefit of Monte-Carlo methods is that they do not necessarily need a model of the environment, but learn from the observed experience. In the case of model-free learning, the action-value function is used for policy estimation rather than the state-value function. The challenge here is that by learning only from the interaction with the environment, we learn only encountered states, and unseen states remain unknown. To make sure that the agent learns about new states, every state-action pair has to have a non-zero chance to be visited. This is a general problem of exploration versus exploitation in reinforcement learning, and in KB-RL we employ an $\epsilon$-greedy policy with respect to the current state-action values to ensure that the exploration will be maintained. The policy is constructed for each action-state pair based on the state-action values following the Normal distribution. Then, most of the time the action is taken based on the constructed policy, however, with probability $\epsilon$ the action is instead selected at random. Section \ref{RLforCR} gives a detailed description of policy construction on the example of the FreeCiv game.

To summarize, the KB-RL approach employs the on-policy model-free Monte Carlo Method  for conflict resolution, averaging over the state-action values and using an $\epsilon$-greedy policy that is maximized on each iteration with regard to the action-value function.

\section{Experiment Setup}

\begin{figure}[t!]
\centering
\includegraphics[width=1.0\linewidth]{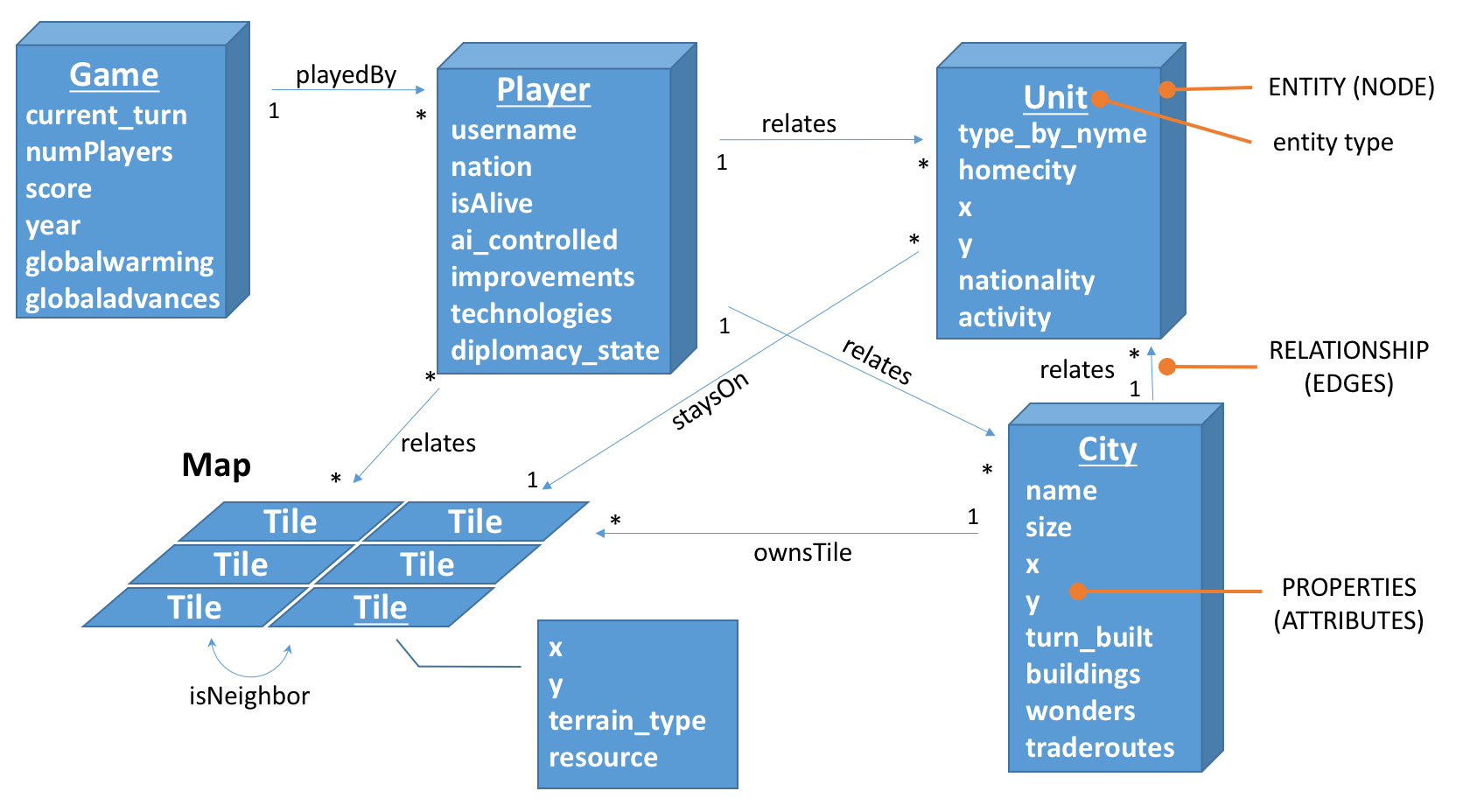}
\caption{Ontology of the FreeCiv game as the representation of the game model for the semantic network within our KB-RL system. For the sake of simplicity, the properties lists are not exhaustive, but rather illustrative.}
\label{fig:ontology}
\end{figure}

This section provides the details of applying the KB-RL method for playing the strategy game FreeCiv. 

\subsection{Game Ontology}
As with any other problem to solve in KB-RL, playing FreeCiv started with outlining the game concepts and their relationships that would be instanced in the semantic network. In other words, we started by defining an ontology as a schema for the semantic graph. FreeCiv is an open-source empire-building strategy game that simulates the history of human civilization. The main concepts in the game are: the game itself, players, units, cities, and the map. Accordingly, the ontology defined these concepts as Entities of the respective types. Figure \ref{fig:ontology} illustrates the ontology developed for the FreeCiv game. The FreeCiv map was constructed from the grid of discrete squares named tiles. Therefore, we introduced to the ontology an Entity of the type Tile, and the map in the semantic network was represented by multiple instances of this type with corresponding relationships amongst them. The attributes of the entity exhibited the properties of the respective concept.

\subsection{Connector and Action Handler}

The purpose of the connector was to recreate a problem environment in the KB-RL semantic graph database and to keep it up-to-date during the entire game. FreeCiv is a turn-based game, thus, all changes in the game happen on the round basis \footnote{except the unit moves that have to be handled on the notification basis.}. The client application saves the state of the game automatically to a dedicated file on each turn. This file holds the full information on the game for a given  player in a given turn. Therefore, the auto-saved files were a perfect source for us to recreate the game state in the graph database. The auto-saved file is a text file that follows a specific structure encoding the characteristics of the game and their values. The Connector monitored the game and on the creation of a new auto-saved file parsed it and saved the changes to the graph database using the REST API. In this manner, the game environment was available to the inference engine to process the task of playing the FreeCiv game. For the sake of fairness, we used only client auto-saved files, as it would be the same for a human player. The universal information about the game that is saved in the server's auto-saved files was unavailable to the agent.

Another service needed for the KB-RL system to play the game was the Action Handler that sent the player's commands to the FreeCiv client. For example, if the rule was instructing the unit to build a city, the command 'unit \textit{id}; press b' would be transmitted by the Action Handler to the game client ( \textit{id} is the numerical identifier of the unit). In the case of FreeCiv, the Action Handler was simply configured to write the commands to a dedicated local file. As FreeCiv is an open-source software, we added the function to the client code that monitored the dedicated file and read the commands as they would have been given by a player through the dialog form. 

\subsection{Game Configuration} \label{gameconfig}

\begin{figure*}[t!]
\centering
\includegraphics[width=1.0\textwidth]{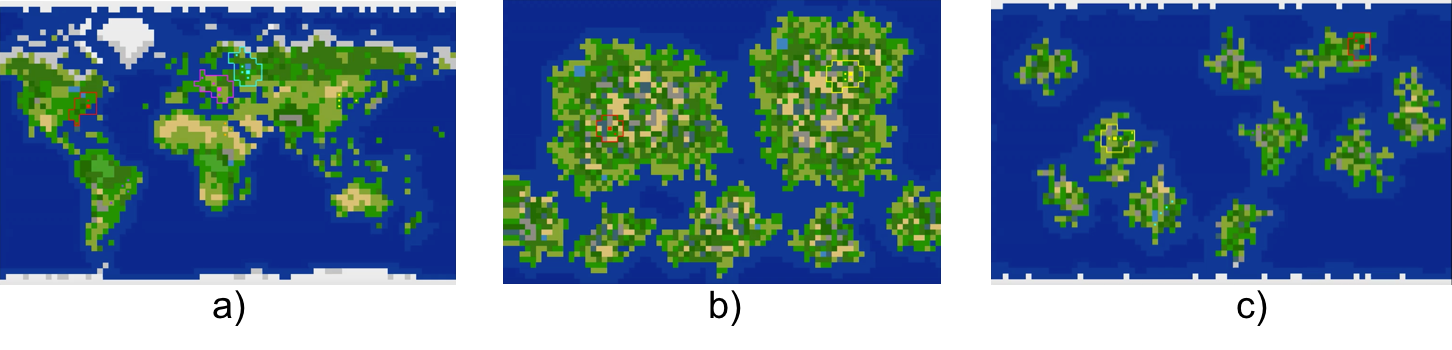}
\caption{Map topology used in the game setups. a) Map topology for Default, Small Islands and USA setups; b) Map topology for Chaos setup; c) Map topology for Medium Islands setup. The color marks the terrain type of the tile.}
\label{fig:maps}
\end{figure*}

FreeCiv is a highly configurable game, down to the specific rules. Players can choose between wide range of settings such as scenario, skill level, number of opponents, a map, nations, etc. We chose to play one of the default scenarios called Earth (classic/small) that had a 80x50 map (4000 tiles) and the 'normal' skill level for all players including AI computer players. We also left the Barbarians on the 'normal' level of difficulty.
For the given scenario, 5 different setups with the map of size 80x50 and fixed starting positions were taken to play the games. The reason for having fixed maps and fixed starting positions rather than random ones was to enforce specific characteristics of the gameplay. Even though FreeCiv offers the game configuration with a random start, after careful consideration, we decided to keep the maps and starting positions fixed. We played several games with random starts and concluded that, due to the randomness, the games were very asymmetric for the different KB-RL agents with different knowledge bases. Therefore, it was very hard to separate the game conditions from the objective evaluation of the knowledge base performance when analyzing the game results. The fixed starting positions were defined by the nation of each player. The 5 setups were as follows:

\begin{itemize}[\noindent]
    \item[] \textbf{Default} - a map with a predesigned classic topology that mimics the Earth (Figure \ref{fig:maps} a). This map was played by 4 players: two KB-RL agents and 2 embedded AI players. The KB-RL agents played for the Roman and Hunnic nations, while the embedded AI played for Aztec and Zulu. On this map, both KB-RL players were within a relatively short distance of each other, not separated by the ocean. Therefore, the players would establish the contact with each other early and could exchange the technologies and be engaged in the trade. Good relationships potentially encouraged collaboration in the Space Colonization race. Thus, the diplomacy was a critical aspect here. 
    \item[] \textbf{USA} - the same as Default map except the nations for KB-RL agents were American and Russian, and there were 4 embedded AI players: Brazilian, Chinese, Arab, and German. In this setup, the KB-RL players were very remote from each other, yet in close distance with the embedded AI players. The closeness of embedded AI players demanded early investment in the defense and military with no opportunity to ally with other KB-RL agent until the sea exploration was developed. Such conditions forced the KB-RL players to follow other decisions than in the Default setup. Diplomacy was not as relevant as previously.
    \item[] \textbf{Small Islands} - the same map as in Default, except the nations for KB-RL agents were British and Japanese, and 3 computer players were Maori, Cuban, and Malagasy. All nations had to start on the very small islands with the land limited to a maximum of 2 cities. That implied the urgency for the sea exploration to access the big land. The focus on production was unavoidable as settlement expansion was not possible. Being out of reach for the opponents conveyed a peaceful start.
    \item[] \textbf{Medium Islands} - the map of 80x50 tiles was customized to create several medium-size islands instead of Earth-like continents (Figure \ref{fig:maps} b). The nations for KB-RL agents were Bahamian and Jamaican, for the embedded AI they were 'Antiguan and Barbudan' and 'Trinidadian and Tobagonian'. The agents had a largely peaceful time to begin the civilization. The size of the islands allowed them to have a comfortable number of cities to build a strong nation and to be well protected from the enemies. However, further expansion was challenging. This map offered a choice for the players in the second phase of the game: develop a limited number of powerful highly populated cities or go overseas and build many small cities capturing more territory.
    \item[] \textbf{Chaos} - a customized map with two huge islands and only two KB-RL players without any embedded AI players (Figure \ref{fig:maps} c). This map offered the players marginal land to settle, however, with relatively poor resources. The remoteness of the islands left the players isolated for a long time before they would discover each other. The absence of computer players allowed the players to focus on the economical development and prosperity of the nation, encouraging space colonization in favor of warfare.
\end{itemize}

Overall, we aimed to configure the maps for encouraging the peaceful win and collaboration between the players rather than hostility. The maps and details of the discussed setups can be examined on the published online histories of the played games \cite{freecivGames}.

\begin{table}
\centering
\footnotesize
\begin{tabular}[t]{|p{3.0cm}|c|c|}
\hline
Expert KI set & Number of rules\\
 \hline
Sentry & 486 \\
Alex & 606 \\
Bernhard Niessl & 605 \\
Tatamo & 724 \\
Martin Kirsch & 613 \\
Mirex & 615 \\
Magnus Wuttke & 604 \\
Bitsquid & 523 \\
Jasper & 577 \\
Suomi & 755 \\
Lemurman & 684 \\  
 \hline
\end{tabular}
\caption{The number of rules in expert KI sets. The expert KI sets are named by the human player usernames.}
\label{tabl:numkis}
\end{table}

\subsection{Knowledge Engineering}

When the semantic network was modelled (ontology), and the communication between the FreeCiv client and the KB-RL agent was set up (Connector, and Action Handler), the next step was to engineer the knowledge. At first, we encoded the game's essential knowledge, such as starting the game, finishing the turn, exploring the land, building and defending cities, producing units, making buildings and wonders, and building a spaceship. These knowledge rules did not contain any strategic decisions for playing the game, but rather the game basics. We refer to this knowledge base as the common knowledge base.

As our next step, we invited experienced players to share their game know-how and to teach us their strategies. The Protocol Analysis method was used in collaboration with the expert players to acquire their knowledge. The experts were asked to play the game and think aloud while doing so. The knowledge engineer was following the game over the shared video stream and could ask the player to comment and explain any aspect of the play. After the game was finished, the knowledge engineer had notes on the player's strategy written down and the history of the game in the form of the auto-saved files, so the game could be reviewed at any later time. We asked each player to play 2 games: on the map of Default setup for 2 different nations that were planned for KB-RL agents.

Overall, we approached 18 people for their expertise in playing FreeCiv. After observing them play, we chose 11 players who showed strong gameplay and confidently won the game against the embedded AI. Their knowledge regarding game strategy was encoded in the additional rule subsets for each player respectively. Joined with the common knowledge base, expert subsets created 11 different expert knowledge bases, for each player's strategy correspondingly. We call them expert KI sets. In general, the strategies differed greatly in such leading decisions as winning by space colonization or by taking over other nations, peaceful or combative behavior, democracy or dictatorship, research or production, control over population, and others. Accordingly, the expert knowledge rules encoded such macro decisions. 

The number of rules in the knowledge base grew gradually throughout the project. The first KB-RL agent that completed the game (despite losing), operated on the knowledge base of around 250 KIs. As the project progressed, the knowledge engineers were continually adding new KIs to the knowledge base. By the time the expert KI sets were implemented, the common knowledge base contained 440 rules. Meanwhile, each expert KI set had various numbers of Knowledge Items to cover the expert knowledge. Table \ref{tabl:numkis} shows the number of KIs for expert KI sets. It has to be noted though that the number of rules in the knowledge base is rather indicative. It can be compared to the lines of code (LOC) metric for software development in the respect that LOC is considered by many to be a very inaccurate metric. For instance, refer to \cite{lineofcode}. For the same reason, the reader must keep in mind that the number of KIs can only be an approximate estimation of the knowledge base complexity and human effort.

\subsection{Tournament phase}

Having 11 expert knowledge bases, we engaged them in combat against each other. In total, there were 550 games played. One game was played by two KB-RL agents provided with two different expert knowledge bases. Each KI set was used in 100 games: 2 games against each of the 10 opponent KI sets on 5 of the maps; these 2 games were played for each of the 2 nations as described in the section \ref{gameconfig}. For example, Alex KI set played once for the Romans and once for the Hunnic on the Default map against 10 other KI sets - 20 games in total. The same was true for 4 other setups that together constituted 100 games. After the tournament, we had collected the records of 1100 games played by KB-RL agent (two clients in each of the 550 games).

The tournament stage of the project provided us with the evaluation of the performance for the created expert knowledge bases. It showed that all of them were strong players to win against computer players and the game could be successfully played with different strategies. Our next concern was to combine the knowledge of all experts into one multi-expert knowledge base and let reinforcement learning support the conflict resolution in the inference process. 

\subsection{Reinforcement Learning for Conflict Resolution} \label{RLforCR}

The FreeCiv's state space for the RL algorithm needed special consideration. The excessive state space is often a challenge for AI problems such as playing strategy games. For example, the chosen FreeCiv configuration has 4000 (80x50) board positions, where each position can have dozens of states. Each tile has a terrain type, can have an improvement like a road, or rail, or irrigation, or special resource; city can be built there, or units can stay on it. The city can have a set of buildings and wonders. The tile can produce different amounts of resources of different types depending on the current government and rates. This list is not exhaustive, but the example illustrates how difficult it would be to account for all possible permutations of the map grid in the FreeCiv game as its state space.
A common practice in playing complex games is adoption of a state reduction technique in order to scale the state space to a manageable number of states \cite{clustering}. In our KB-RL approach, we applied clustering to segment the game's state space into a finite number of clusters.

The clustering dataset was created based on 1100 game histories from the tournament phase with the selected game features. We started with the analysis of the game features and their correlation with the won/lost outcome of the games. Features that showed the strongest relationships between their values and the result of the game were added to the dataset. Overall, 33 features were selected: game score; population size; rates for tax, luxury, and science; amount of generated resources per turn such as gold, production, science; accumulated natural resources such as gold, production, science; number of explored, owned, and owned by enemy tiles; number of ocean tiles; sum of defense and attack points for all the units; diplomacy state; number of players; maximum, minimum and average of the number of cities in the dataset; maximum, minimum and average of the game score throughout the dataset; maximum, minimum and average of the number of learned technologies in the dataset; nation; government type; number of learned global technologies; number of learned technologies by given player; and learned technologies. Weights were applied to the feature vector based on how strong the correlation between the feature values and the won/lost output was. Essentially, the game score, population size, and the number of learned technologies (both, global and national) were weighted the highest, while the type of learned technologies influenced the winning/losing rate the least, and was therefore weighted lowest. Other features received moderate weights to achieve the best clustering accuracy. The collected dataset had 386895 entries.

The k-means algorithm was used to conduct the clustering on the normalized dataset. Particularly Lloyd's algorithm with a maximum of 300 iterations was applied, and 185 clusters were defined as a result of experimenting with various possible numbers of clusters. 185 clusters represented the generalized game states with respect to the feature parameters, and they composed the state space for the FreeCiv game in KB-RL.
To map the game situation to the cluster during the game play, the feature vector was constructed from the current parameters and the distance to the cluster centers was calculated for every turn. The closest cluster (minimal distance) was assigned to be the game state in the current turn. From now on, when using the term 'state', we will refer to the cluster that was assigned to the game in the given turn.

Usually, the game remained in one cluster for more than one turn. We defined the \textit{cluster turn} as the mean of all turns that were assigned to the given cluster. Thus, if the game had been in the cluster $C$ in all turns from $a$ to $b$, the cluster turn was then $T_C=\sum_{i=a}^{b}i/(b-a+1)$. For instance, the game had been in the cluster $C$ in turns from 7 to 19, then $T_C=13$. Across multiple games, the cluster turn was averaged again, so during training every cluster was given a number, referred as \textit{cluster turn}, indicating the mean turn for the game to be in this state. The cluster turn was used to determine the state return with respect to the defined goal.

Foremost, for learning the optimal behavior the RL needed an outlined goal in regards to its state in the environment. In the FreeCiv game, it is not enough to just win the game, but the winner has to race with other players to accomplish the game before other opponents. Intuitively, the players seek to minimize the number of rounds it takes to win the game, making shorter play time an indication of the player's proficiency and competence. The same conclusion was drawn from the analysis of 1100 games derived from the tournament phase. The longer it took for the player to learn technologies, build a spacecraft and reach Alpha Centauri, the less likely that the game was won. Therefore, the reward function for the RL agent was chosen to be based on the number of turns the game lasted, with the objective of minimizing the game rounds to win. The defined reward function returned -1 for each turn played in the game, and we did not discount the return. Consequently, for each winning episode, its \textit{return} was defined as $G=-N$, where $N$ is the last turn of the game. Starting from state $s$, the return for the state was defined as $G_s=-(N-t)$, where $t$ is the cluster turn of the state $s$. In essence, $G_s$ indicated the expected number of turns that agent would need from the current state to finish the game. According to the Monte Carlo method, the state-action function for the given policy was then derived from averaging the sampled returns for the given state-action pairs.

There are two possible outcomes for the agent: winning or losing. If the game was won, the return was defined as discussed above. If the game was lost, however, the return had to indicate that the result of the episode was not desired. For the lost games, we also distinguished between the way they were lost: either the opponent reached Alpha Centauri first, or the player's nation was destroyed in war. We preferred to teach the agent a peaceful course of the game. Losing in space colonization was punished less than being destroyed because it was interpreted as the player being sufficiently strong to withstand the opponents' attacks and only lacking the time to reach the remote star. Therefore, for the first case, the episode's return was given as $G=-N*2$, while for the later, the return was set as $G=1000-N$ (the earlier the player was destroyed, the lower was the return). For the state $s$ the return was calculated as $G_s=-(N*2-t)$ or $G_s=-(1000-N+t)$ respectively.

The action space for the trained agent was composed of all the Knowledge Items of the multi-expert knowledge base, so that every KI was considered as an action $a\in A$. In every situation when the inference engine encountered a conflict, the learned policy was applied to the conflict set and one selected KI was executed. After every episode, the state-action values were calculated, and the policy was improved based on the new values for the next episode. 

The RL algorithm for playing FreeCiv game used the stochastic policy that followed the Normal distribution. Normal distribution is defined by two parameters, the mean $\mu(s)$ and the standard deviation $\sigma(s)$. The parameters $\mu$ and $\sigma$ were constructed after each episode for each state-action pair based on the state-action values and the expected return of the given state.

Firstly, for every cluster the state-action values were scaled between 0 and 1 following min-max normalization. The normalized state-action values were then taken as the mean $\mu_s^a$ parameter for the action $a$ in the state $s$. The standard deviation $\sigma_s^a$ was calculated as in equation \ref{eq:deviation}
\begin{equation}
    \sigma_s^a=\sqrt{\frac{\sum{(G_s-\overline{G}_s)^2}}{n_a}}
    \label{eq:deviation}
\end{equation}
where $G_s$ is the return of the specific cluster $s$ in which the action $a$ occurred, and $\overline{G}_s$ is the expected return of the cluster $s$.

\begin{figure}[t!]
\centering
\includegraphics[width=1.0\linewidth]{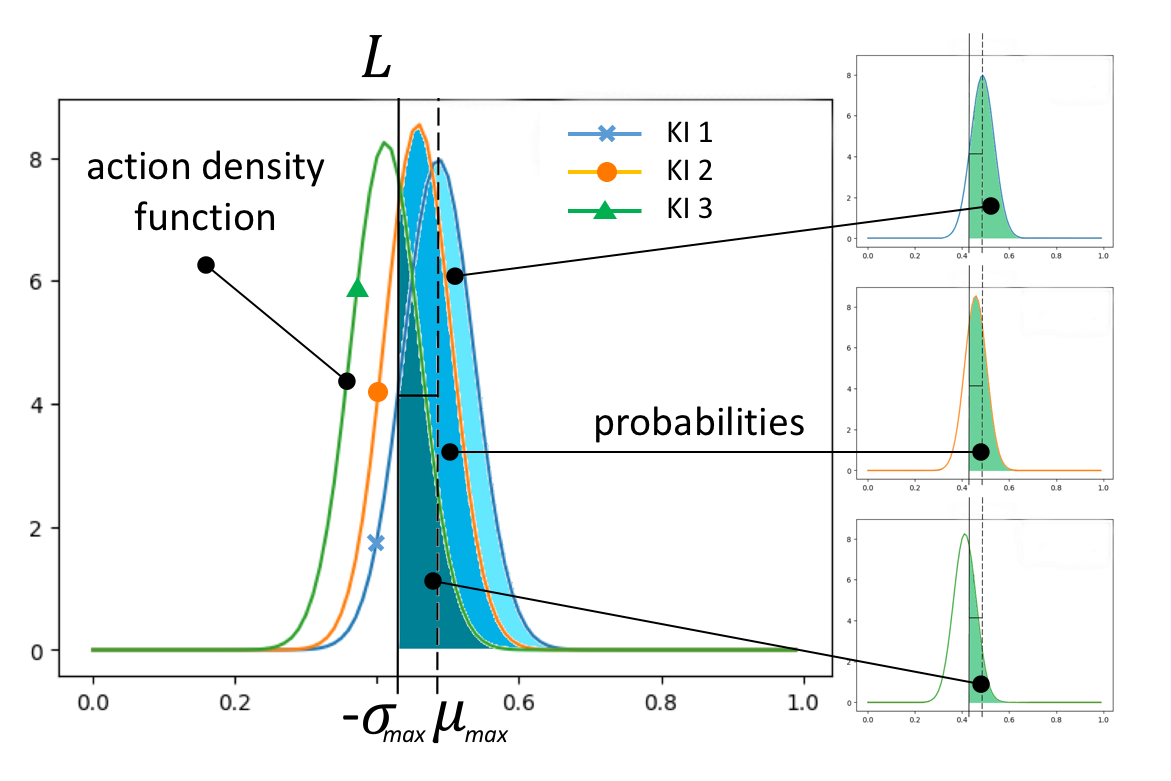}
\caption{The probabilities for the KIs (actions) of the conflict set derived as an area under the curve on the right to the limit line L. The KI 1 has the highest action-value, thus $\mu$ and $\sigma$ parameters calculated for KI 1 are taken to form the limit $L=\mu_{max}-\sigma_{max}$.}
\label{fig:probabilities}
\end{figure}

\begin{figure}[t!]
\centering
\includegraphics[width=0.7\linewidth]{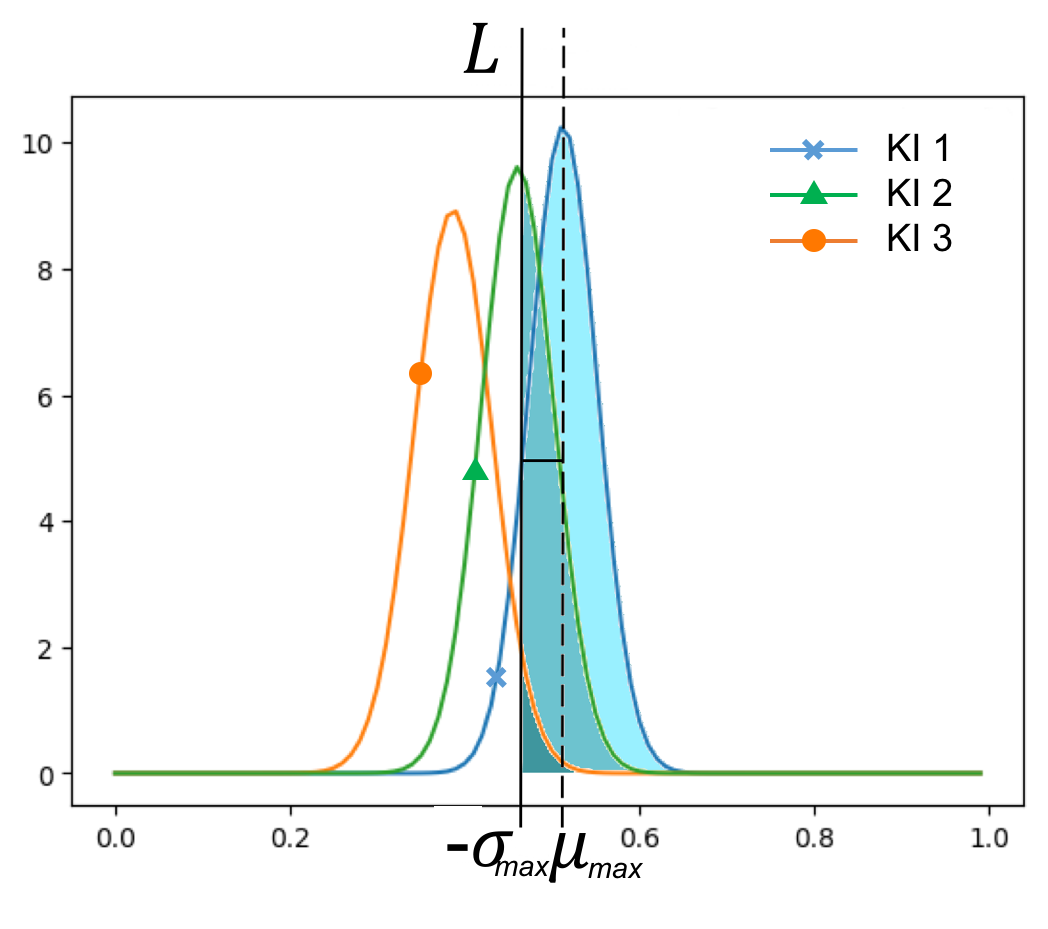}
\caption{The probabilities later in the training. As it can be seen from the picture, the probability of the KI 1 increased significantly as a result of more experience and high return. On the contrary, the probability for the KI 3 diminished due to low return.}
\label{fig:laterprobabilities}
\end{figure}

While the parameters $\mu$ and $\sigma$ were calculated for every state-action pair after each episode, the probabilities for choosing action $a$ in state $s$ were calculated at the time when the inference engine encountered a conflict and the conflict set underwent resolution. Then, the probabilities were calculated for each action of the conflict set, and the action was selected according to these probabilities. The probabilities were calculated as followed. The probability density function for each KI of the conflict set had been drawn with the parameters $\mu_s^a$ and $\sigma_s^a$, where $a$ is a KI and $s$ is the cluster. The highest value of $\mu$ ($\mu_{max}$) and the standard deviation of this action ($\sigma_{max}$) were taken to form a limit $L=\mu_{max}-\sigma_{max}$. The probabilities were then acquired as the normalized areas under the curve on the right from the limit $L$ line. Figure \ref{fig:probabilities} illustrates this algorithm in the case of three KIs in the conflict set. 
While $\mu$ indicated that for the higher values there would be a better outcome for the given action, the parameter $\sigma$ can be interpreted as a confidence in the acquired state-action values. The more times action $a$ was tried, the closer $\sigma$ was to the value $\mu$. The closer $\sigma$ was to $\mu$, the more separated the bell curves were for the actions, and consequently, a higher probability was given to the best action, and smaller probabilities remained for other actions, as shown in Figure \ref{fig:laterprobabilities}.

We ran the training algorithm for 600 episodes. One full game lasted approximately 8-12 hours, therefore presenting a challenge to run extra episodes. During the training phase, the game was set up with 4 players where one was a KB-RL agent with the multi-expert knowledge base, one KB-RL agent was taken either with the multi-expert knowledge base or with one of the expert knowledge bases, and 2 embedded AI players. The training phase was terminated after 600 games and we set the trained agent to play against each of expert knowledge bases to explore the result. The trained agent played 2 games in each of the 5 setups against each expert knowledge base as it was arranged in the tournament phase discussed above. The next section discusses the results of this contest and of the overall experiment.

\section{RESULTS}

\begin{figure}[t!]
\centering
\includegraphics[width=1.0\linewidth]{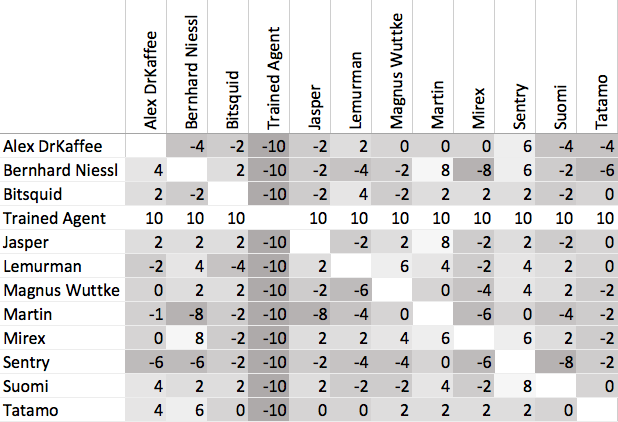}
\caption{The result table for the tournament between expert rule sets and the trained agent. The expert rule sets are named by the username of the human player whose strategy was encoded in the particular rule set. Each cell has to be read as follows. The player in the row label played against the player in the column label 10 games. The number shows the difference in the number of wins for the player given in a row label. For example, Alex DrKaffee played against Bernhard Niessl. Alex DrKaffee won 3 games, while Bernhard Niessl won 7 games. Alex DrKaffee' advantage is -4, while Bernhard Niessl holds the advantage 4. The table is almost symmetrical except the results for Martin versus Alex DrKaffee: one of their games was lost by both KB-RL agents.}
\label{fig:tablewithhiro}
\end{figure}

\begin{figure}[t!]
\centering
\includegraphics[width=0.9\linewidth]{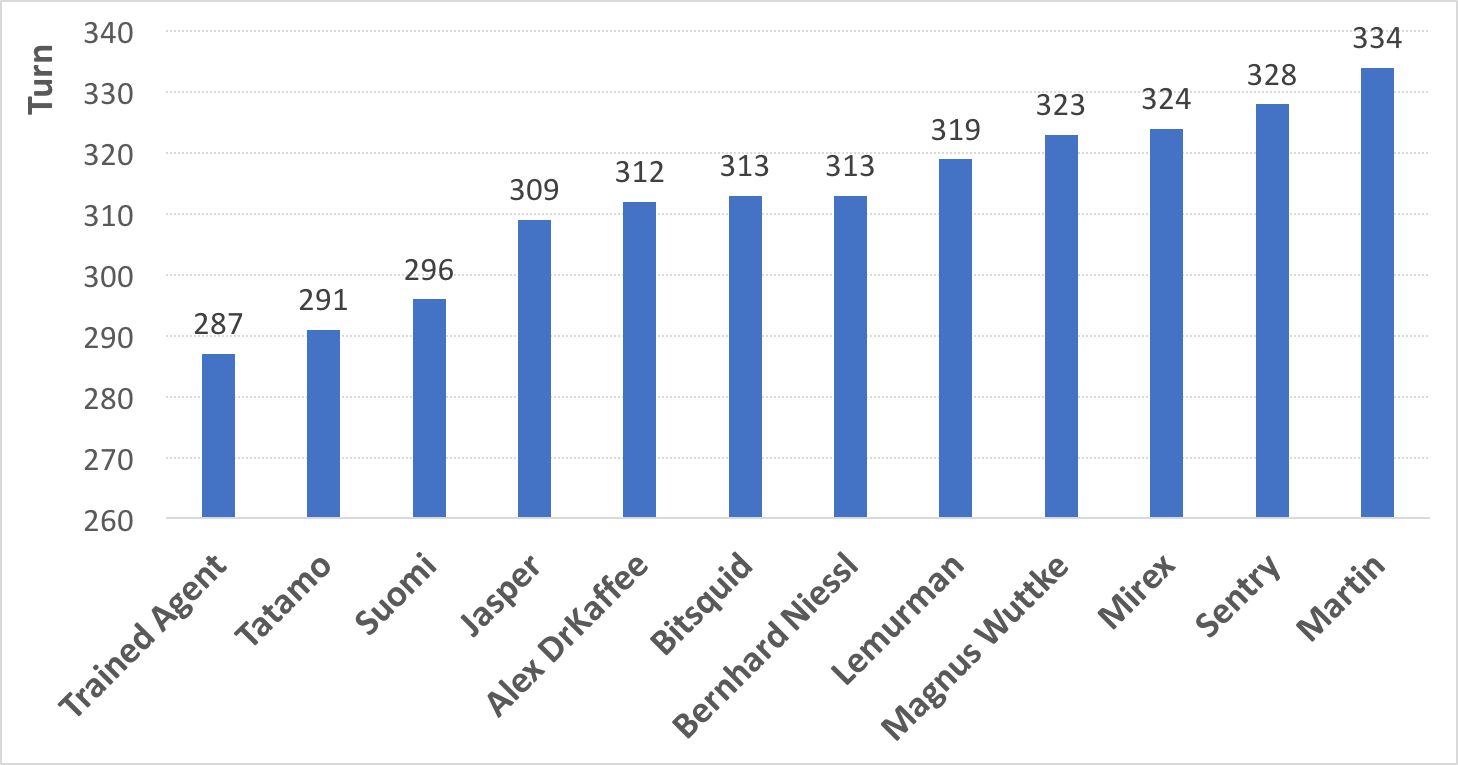}
\caption{Average number of turns that the agents needed to win the game. The results are calculated for the games played in the tournament.}
\label{fig:avgTurn}
\end{figure}

\begin{figure}[t!]
\centering
\includegraphics[width=0.9\linewidth]{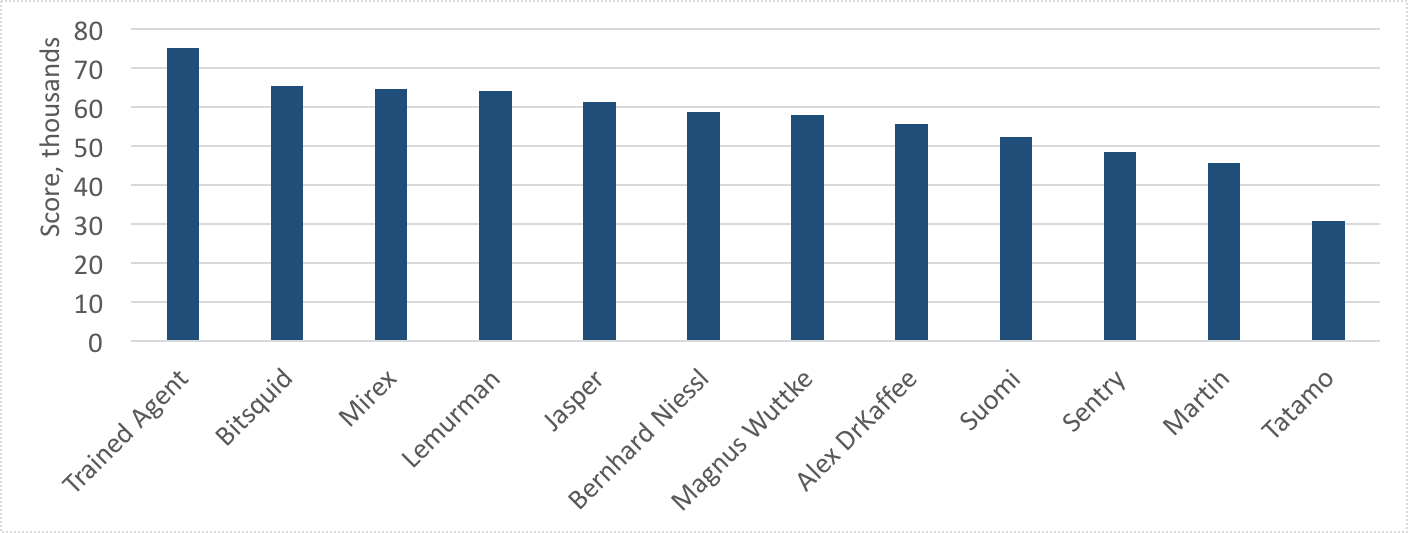}
\caption{Summed final scores for the agents of the tournament. As all agents played the same number of games, the averaged final score would reveal the same trend.}
\label{fig:score}
\end{figure}

\begin{figure}[t!]
\centering
\includegraphics[width=0.9\linewidth]{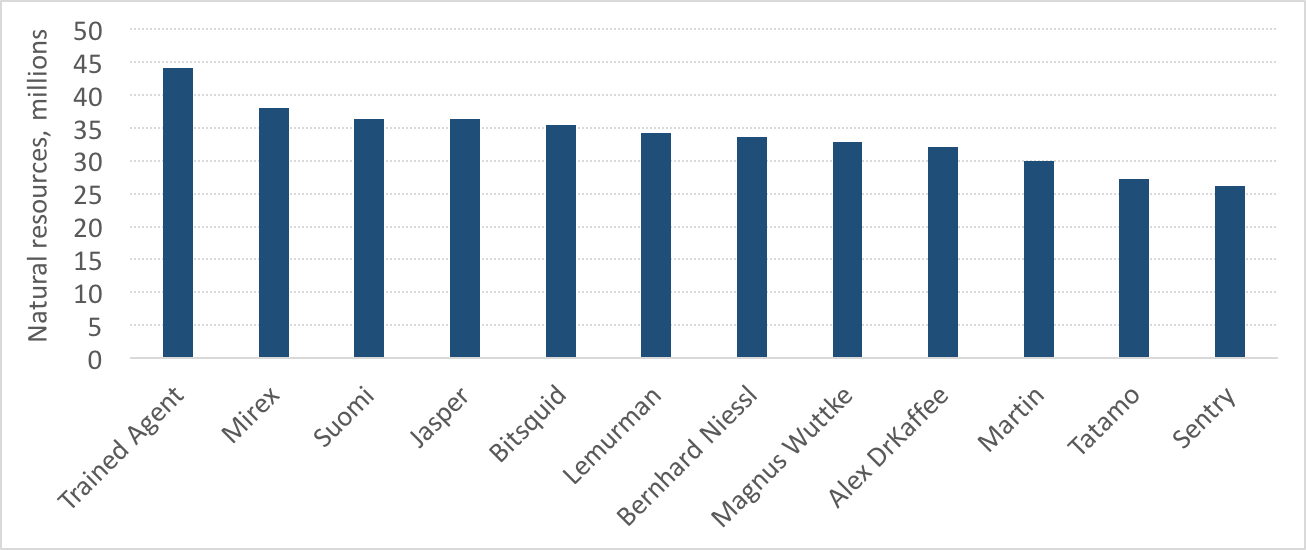}
\caption{Generated natural resources over the full game for the agents of the tournament. As all agents played the same number of games, the averaged numbers for the generated natural resources would reveal the same trend.}
\label{fig:output}
\end{figure}

\begin{figure}[t!]
\centering
\includegraphics[width=0.9\linewidth]{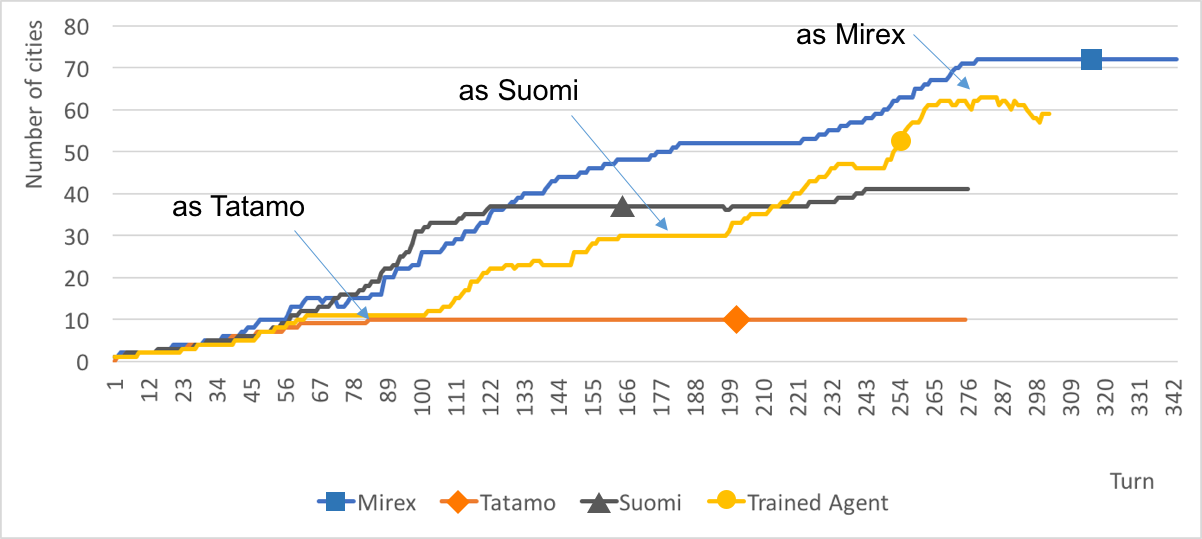}
\caption{Number of cities for the trained agent and the expert rule sets on the example of single games. The picture illustrates how the trained agent adopts the strategies of different experts in one game.}
\label{fig:citiesVsTrun}
\end{figure}

\begin{figure}[t!]
\centering
\includegraphics[width=0.9\linewidth]{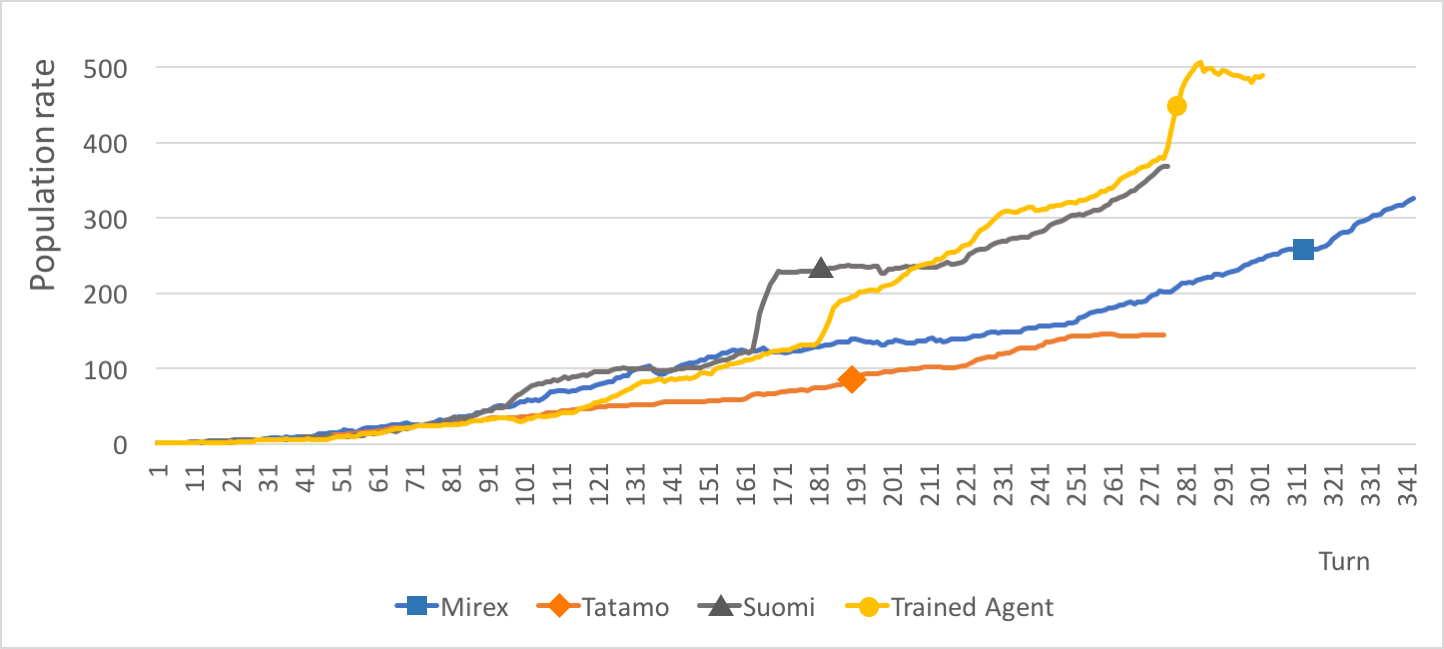}
\caption{The population rate for the trained agent and the expert agents that corresponds to the data in Figure \ref{fig:citiesVsTrun}. The trained agent was able to grow the population above other expert players.}
\label{fig:populationVsTurn}
\end{figure}

The described experiment shows that the KB-RL approach can be successfully applied to solve complex tasks such as playing a strategy game, particularly FreeCiv. At first, with limited and imperfect knowledge, and without reinforcement learning,  the agent was able to finish all games, though mostly lost. Such an agent can be compared to a human player at the beginner level. Its knowledge was elicited from our team members who are software engineers rather than well-experienced players.
Without sophisticated strategic decisions inserted, the results of playing with the common knowledge base were poor.

Next, by adding expert knowledge to the common knowledge base, the solution was extended to 11 different solutions, where each of them represented a particular strategy of a human player. These solutions demonstrated significantly stronger play in contrast to the agent with the common knowledge base. Each of the expert knowledge bases could already win against embedded AI players in the majority of the games. Figure \ref{fig:tablewithhiro} (all players except the trained agent) shows the results table of the combat between expert knowledge bases. There was only one game (Martin versus Alex DrKaffee in the USA setup) won by the computer player, while the rest of the games was won by one of the KB-RL agents equipped with the particular expert knowledge base. 

While the above results were achieved with just the knowledge-based approach, the main result of the discussed KB-RL method derives from reinforcement learning. Reinforcement learning was deployed as a conflict resolution strategy to be able to combine the knowledge of multiple experts in one knowledge base.
In alignment with the outlined goal, the KB-RL system learned to win the game on average in fewer turns in comparison to the agents with the expert knowledge bases. As it can be seen in Figure \ref{fig:avgTurn}, the trained agent needed on average 287 turns to win, while for the expert knowledge bases the best average number of turns was 291 for the Tatamo expert knowledge base. This advantage allowed the trained agent to win every game played against expert knowledge bases as shown in Figure \ref{fig:tablewithhiro}.

The dominance of the trained agent in respect to the expert knowledge bases can also be seen in other aspects of the game, such as the amount of generated resources and the game score (Figures \ref{fig:score} and \ref{fig:output}).

Analyzing the results of the trained agent in contrast to the expert agents, it was found that the trained agent picked up the fragments from different expert knowledge bases to make up its own strategy. For instance, the Tatamo agent kept the number of cities strictly under 10 to maintain control over the citizens. The Suomi agent had a rule to stop building cities after the first 30-40 cities, while other agents did not control the number of cities. The trained agent combined the control of the city number into a new strategy: in the first phase of the game, the agent followed Tatamo strategy and had 11 cities, however later it switched to the Suomi strategy and increased the number of cities to above 30, and again changed its strategy to unrestricted settling in the final stage of the game. Figures \ref{fig:citiesVsTrun} and \ref{fig:populationVsTurn} illustrate the given pattern on the exemplified games.

\section{DISCUSSION}

This paper describes the KB-RL approach as a knowledge-based method combined with reinforcement learning in order to deliver a system that leverages the knowledge of multiple experts and learns to optimize the problem solution with respect to the defined goal. RL is employed as a conflict resolution strategy for the multi-expert knowledge base with excessive knowledge for a particular problem solution. The method is demonstrated by the example of playing a complex strategy game such as FreeCiv. The knowledge and skills of several game experts were combined into one knowledge base and set up to play incrementally, learning to win the game in the minimal number of turns. The results show that RL can improve the performance of the agent by learning to recombine the elements of single strategies into a new solution stimulated by the outlined objective.

The proposed approach leaves much room for future work and further research. Overall, the described experiment supports the idea of bringing together different AI approaches for more intelligent and better automated systems that can utilize human knowledge and learn from its own experience in complex problem solving. 


%
%


\bibliographystyle{plain} 
\bibliography{bibliography}   

\end{document}